\documentclass[runningheads]{llncs}

\usepackage{cite}

\newcommand{\adj}{\mathbf{A}}
\newcommand{\perm}{\mathbf{P}}
\newcommand{\ortho}{\mathbf{O}}
\newcommand{\trans}{\mathbf{t}}
\newcommand{\fea}{\mathbf{F}}
\newcommand{\feavec}{\mathbf{f}}
\newcommand{\emb}{\mathbf{H}}
\newcommand{\embvec}{\mathbf{h}}

\newcommand{\messvec}{\mathbf{m}}
\newcommand{\pos}{\mathbf{X}}
\newcommand{\posvec}{\mathbf{x}}

\usepackage{amsfonts}
\usepackage{amsmath}
\usepackage{comment}
\usepackage{pdfpages}

\usepackage{graphicx}

\usepackage{caption}
\usepackage{siunitx,booktabs}
\usepackage{hyperref}
\hypersetup{
    colorlinks=true
}

\usepackage{graphicx}

\usepackage{hyperref}

\makeatletter
\newcommand{\printfnsymbol}[1]{%
  \textsuperscript{\@fnsymbol{#1}}%
}
\makeatother
\DeclareMathOperator{\iou}{IoU}
\DeclareMathOperator{\dice}{Dice}

\begin{document}
\title{Utility of Equivariant Message Passing in Cortical Mesh Segmentation}

\author{
    Dániel Unyi\inst{1} \and
    Ferdinando Insalata\inst{2} \and \\
    Petar Veličković \inst{3}\thanks{equal contribution} \and
    Bálint Gyires-Tóth \inst{1}\printfnsymbol{1}
}

\institute{
    Department of Telecommunications and Media Informatics\\ 
    Budapest University of Technology and Economics\\ Műegyetem rkp. 3., H-1111 Budapest, Hungary\\
    \email{\{unyi.daniel,toth.b\}@tmit.bme.hu}
    \and
    Department of Mathematics, Imperial College London\\
    London SW7 2AZ, United Kingdom\\
    \email{f.insalata17@imperial.ac.uk}    
    \and
    DeepMind\\
    \email{petarv@deepmind.com}    
}


\authorrunning{D. Unyi et al.}
%

%
\maketitle              
\begin{abstract}
The automated segmentation of cortical areas has been a long-standing challenge in medical image analysis. The complex geometry of the cortex is commonly represented as a polygon mesh, whose segmentation can be addressed by graph-based learning methods. When cortical meshes are misaligned across subjects, current methods produce significantly worse segmentation results, limiting their ability to handle multi-domain data. In this paper, we investigate the utility of E(n)-equivariant graph neural networks (EGNNs), comparing their performance against plain graph neural networks (GNNs). Our evaluation shows that GNNs outperform EGNNs on aligned meshes, due to their ability to leverage the presence of a global coordinate system. On misaligned meshes, the performance of plain GNNs drop considerably, while E(n)-equivariant message passing maintains the same segmentation results. The best results can also be obtained by using plain GNNs on realigned data (co-registered meshes in a global coordinate system).


\end{abstract}

\section{Introduction}

It has long been recognized that machine learning is an important technique in medical data analysis. With advances in deep learning, state-of-the-art solutions are even able to outperform medical professionals on certain datasets~\cite{liu2019comparison}. fMRI allows the non-invasive and non-radioactive examination of the brain, and there has been much interest in the application of deep learning to fMRI scans~\cite{dl2018fmri,dl2018fmri2,dl2020fmri}. An important area of scan analysis focuses on segmentation, i.e. classifying unstructured 2-, 3-, and 4-dimensional scans by comparing certain statistical properties. Manual segmentation is a labor-intensive process that requires highly-trained experts, hence the interest in automating it. Automatic segmentation assists doctors to identify normal and abnormal regions. The difficulty stems from the variable nature of human brains, scanning methods, and environmental conditions. The scans can be segmented in 2-dimensions as images, in 3-dimensions as meshes (or point clouds), and in 4-dimensions that also includes temporal information.

Historically, the main approach in automatic segmentation was to apply computer vision techniques. Within deep learning, mainly convolutional neural networks (CNNs) were applied~\cite{dl2018fmri2,dl2020fmri}. 3D CNNs are able to work on volumetric data, and were successfully utilized for segmentation tasks in the U-Net structure~\cite{unet,2018s3dunet}. Lately, with the rise of the transformer architecture~\cite{2017transformer}, such methods were also applied for segmentation tasks separately or combined with CNNs~\cite{2021transformertransunet,2021transformerswinunet,2021transformercnnMedical}. 

With the marching cubes algorithm, the brain can be reconstructed as a 3D mesh by combining multiple MRI scans taken from the parallel brain slices of a patient~\cite{1987marchingcubes}. 
Reconstruction methods involve several preprocessing steps (e.g. thresholds are set based on the data, sometimes subjectively) that introduce minor or major distortions to the resulting mesh. Reconstruction quality is also heavily influenced by the resolution of MRI scans. When machine learning techniques are applied to 3D meshes -- which are reconstructed from 2D images -- these constraints create an irreducible error barrier. In spite of these constraints, reconstructing 3D meshes can still provide a greater amount of spatial information than analyzing 2D images separately.

The cerebral cortex is a sheet of neural tissue located on the outer surface of the brain. 
A large number of neurons live within its folds and grooves (up to 16 billion neurons in humans), which facilitate the processing of large amounts of information. Cortical meshes have emerged as a popular way of representing its complex geometry, being a valuable tool for studying patterns in healthy brains as well as the structural and functional abnormalities that accompany pathological conditions. Complex cognitive processes, including sensory, motory and association, involve distinct areas of the cortex. The task of segmenting these areas is therefore of great scientific and medical interest.


In this work, we investigate the segmentation performance of neural network architectures, that (i) processes each node of the cortical meshes separately with a multilayer perceptron (ii) considers the underlying geometry, i.e. the edge structure of the cortical meshes with a graph neural network~\cite{gnn}, (iii) considers the underlying geometry, and is equivariant to isometric transformations with E(n)-equivariant graph neural networks~\cite{egnn}. 

The main contributions of this paper are the following:
\begin{itemize}
    \item We are the first to investigate the utility of E(n)-equivariant graph neural networks (EGNNs) in cortical mesh segmentation.
    \item We evaluate the segmentation performance of EGNNs against plain graph neural networks (GNNs), both on aligned and misaligned cortical meshes.
    \item We explain why GNNs are better choice for segmentation than EGNNs when the cortical meshes are aligned or can be realigned in a global coordinate system.
\end{itemize}

\section{Background}

\subsection{Graph Neural Networks}

Graph Neural Networks (GNNs) are neural networks which operate on graph-structured data~\cite{bruna,defferrard,kipf,velickovic}. Let \( \mathcal{G = (V, E)} \) be a graph with \(N\) nodes, adjacency matrix \(\adj \in \mathbb{R}^{N \times N} \) and node embedding matrix \(\emb = (\embvec_0, \embvec_1, ... \embvec_N) \in \mathbb{R}^{N \times D} \).
Since nodes can be re-indexed arbitrarily, a common property of GNN layers is permutation equivariance:
\[ f(\perm \emb, \perm \adj \perm^T) = \perm (f(\emb, \adj)) \]
where \(f\) is a GNN layer and \(\perm \in \mathbb{R}^{N \times N} \) is an arbitrary permutation matrix.
Such functions can be constructed in numerous ways, and GNN layer design is a remarkably active research area~\cite{gnn_survey,geometric}. One of the most expressive GNN layer is the message passing layer, proposed by Gilmer et al. for quantum chemical applications~\cite{gnn}. The l-th layer is constructed as follows:
\begin{enumerate}
    \item Concatenate the node embeddings along the edges, and transform the resulting edge embeddings using a small MLP \(\phi_e\):
    \[ \messvec_{ij}^l = \phi_e(\embvec_i^l, \embvec_j^l) \]
    \item Sum up the edge embeddings in each neighbourhood:
    \[ \messvec_{i}^l = \sum_{j \in \mathcal{N}(i)}\messvec_{ij}^l \]
    \item Concatenate the updated node embeddings to the original ones, and transform the resulting node embeddings using a small MLP \(\phi_h\):
    \[ \embvec_{i}^{l+1} = \phi_h(\embvec_i^l, \messvec_i^l) \]
\end{enumerate}

\subsection{E(n) Equivariance}
In MRI-based mesh segmentation and many other applications (e.g. point clouds \cite{point cloud}, 3D molecular structures~\cite{3D molecular structure}, or N-body simulations~\cite{n-body simulation}), graphs are embedded into 3D Euclidean space. It means that beyond node features \(\fea = (\feavec_0, \feavec_1, ... \feavec_N) \in \mathbb{R}^{N \times F} \), the node coordinates \( \pos = (\posvec_0, \posvec_1, ..., \posvec_N) \in \mathbb{R}^{N \times 3} \) are also available. Now one can not only choose the arbitrary permutation of node indices, but also an arbitrary basis which the node coordinates are represented on. E(n)-equivariant GNNs turned out to be more effective in these applications~\cite{geometric,e3,se3}, and recently, Satorras et al. proposed a very simple and effective E(n)-equivariant GNN (EGNN) based on message passing~\cite{egnn}. The l-th layer is constructed as follows:
\begin{enumerate}
    \item Concatenate the node embeddings along the edges as well as the node distances, and transform the resulting edge embeddings using a small MLP \(\phi_e\):
    \[ \messvec_{ij}^l = \phi_e(\embvec_i^l, \embvec_j^l, ||\posvec_{i}^l - \posvec_{j}^l||) \]
    \item Sum up the edge embeddings in each neighbourhood:
    \[ \messvec_{i}^l = \sum_{j \in \mathcal{N}(i)}{\messvec_{ij}^l} \]
    \item Concatenate the updated node embeddings to the original ones, and transform the resulting node embeddings using a small MLP \(\phi_h\):
    \[ \embvec_{i}^{l+1} = \phi_h(\embvec_i^l, \messvec_i^l) \]
    \item Update the node coordinates, using a small MLP \(\phi_x\):
    \[ \posvec_{i}^{l+1} = \posvec_i^l + \frac{1}{|\mathcal{N}(i)|} \sum_{j \in \mathcal{N}(i)} {\phi_x(\messvec_{ij})(\posvec_i^l - \posvec_j^l)} \]
\end{enumerate}
The elements of the E(n) group are orthogonal transformations (i.e. rotations and reflections) and translations, collectively called isometric transformations as they preserve the length of the transformed vectors. Let \(\ortho\) be an orthogonal matrix, \(\trans\) a translation vector, and \(f\) an EGNN layer. Since node embeddings depend only on the distances between nodes, their transformation rule is E(n)-invariant:
\[ f(\ortho \emb + \trans, \adj) = f(\emb, \adj) \]
Node coordinates are also updated in each layer, such that their transformation rule is E(n)-equivariant:
\[ f(\ortho \pos + \trans, \adj) = \ortho f(\pos, \adj) + \trans \]

\subsection{Point Cloud Registration}

Assume we measure two point clouds \( \pos \in \mathbb{R}^{N \times 3} \) and \( \pos' \in \mathbb{R}^{N \times 3} \) which are identical up to an isometric transformation. The goal is to estimate the isometric transformation, by minimizing the error function
\[ \sum_{i=1}^N || \posvec_i' - (\ortho \posvec_i + \trans) ||^2 \]
According to Arun et al.~\cite{point_cloud}, this problem can be solved exactly:
\begin{enumerate}
\item Compute the centre of mass of each point cloud:
\[ \mu = \frac{1}{N} \sum_{i=1}^N \posvec_i \quad \text{and} \quad \mu' = \frac{1}{N} \sum_{i=1}^N \posvec_i' \]
\item Compute the point cloud matrix \(\mathbf{W}\) and its singular value decomposition:
\[ \mathbf{W}  = \sum_{i=1}^N (\posvec_i - \mu) (\posvec_i' - \mu') ^T = \mathbf{U} \mathbf{\Sigma} \mathbf{V}^T \]
\item Express the isometric transformation as the following orthogonal matrix and translation vector:
\[\ortho = \mathbf{U}\mathbf{V}^T \quad \text{and} \quad \trans = \mu' - \ortho \mu \]
\end{enumerate}

\subsection{Related Work}

Deep learning methods have been used in several previous studies to segment cortical areas, including MLPs~\cite{mlp}, CNNs~\cite{cnn_1,cnn_2}, spherical CNNs~\cite{spherical_cnn_1,spherical_cnn_2}, mesh CNNs~\cite{multi_cnn}, and GNNs~\cite{multi_cnn,gnn_1,cucurull}. MRI scans are very often transformed into meshes before segmentation; and in many cases, meshes are further simplified by mapping their surfaces to planes (for CNNs) or spheres (for spherical CNNs), which may lead to the loss of potentially important information. Mesh-based representations clearly have the advantage of carrying lots of information about the local and global geometric relationships of the cortical surface. Due to their immediate applicability on these irregular surfaces, mesh CNNs and GNNs are more successful in this task.

Cucurull et al. were the first to apply GNNs to cortical mesh segmentation, focusing on Broca's area~\cite{cucurull}. They demonstrated that by incorporating the edge structure of the meshes, GNNs~\cite{defferrard,kipf,velickovic} significantly improve on previous state-of-the-art methods. The node coordinates were also concatenated to the node features, which further improved the segmentation accuracy.

Rotational equivariance was studied in a recent work by Fawaz et al.~\cite{multi_cnn}. They mapped the cortical surfaces to a sphere, and tested various geometric deep learning methods in two tasks: cortical segmentation and neurodevelopmental phenotype prediction. They found that rotational equivariance is less important than filter expressivity or the method of pooling, provided the whole dataset was pre-aligned. Non-equivariant filters showed deteriorated performance when the test data was rotated.

The importance of cortical mesh alignment was also recognized by Gopinath et al.~\cite{alignment}. Using adversarial training, they trained two models: a segmentator GNN and a discriminator GNN. The segmentator GNN was trained to segment both aligned and unaligned meshes, and the discriminator GNN was trained to predict whether the segmentation result comes from an aligned or an unaligned mesh. The discriminator loss was minimized while the discriminator parameters were updated, but maximized while the segmentator parameters were updated, hence the segmentator was forced to segment the aligned and the unaligned meshes similarly. In this case, instead of the structure of the network, the network parameters carry the alignment invariance.


\section{Methodology}

The goal of the proposed work is to investigate whether and how co-registering in a global coordinate system (GCS) supports the segmentation of cortical meshes. For this purpose, we evaluate the performance of the E(n)-equivariant graph neural network~\cite{egnn} against its non-equivariant version~\cite{gnn} in a series of experiments, in which:
\begin{enumerate}
    \item We use the GCS defined by the initially co-registered meshes;
    \item We drop this GCS by applying random isometric transformations to each of the meshes;
    \item We define another GCS by estimating the isometric transformations between the meshes and an arbitrarily chosen reference mesh.
\end{enumerate}
In all cases, we provide brief explanations to clarify the results achieved by the networks. Code to reproduce the results is available on GitHub\footnote{\url{https://github.com/daniel-unyi-42/Equivariant-Cortical-Mesh-Segmentation}}. 

\begin{figure}[!t]
  \centering
  \includegraphics[width=0.95\linewidth]{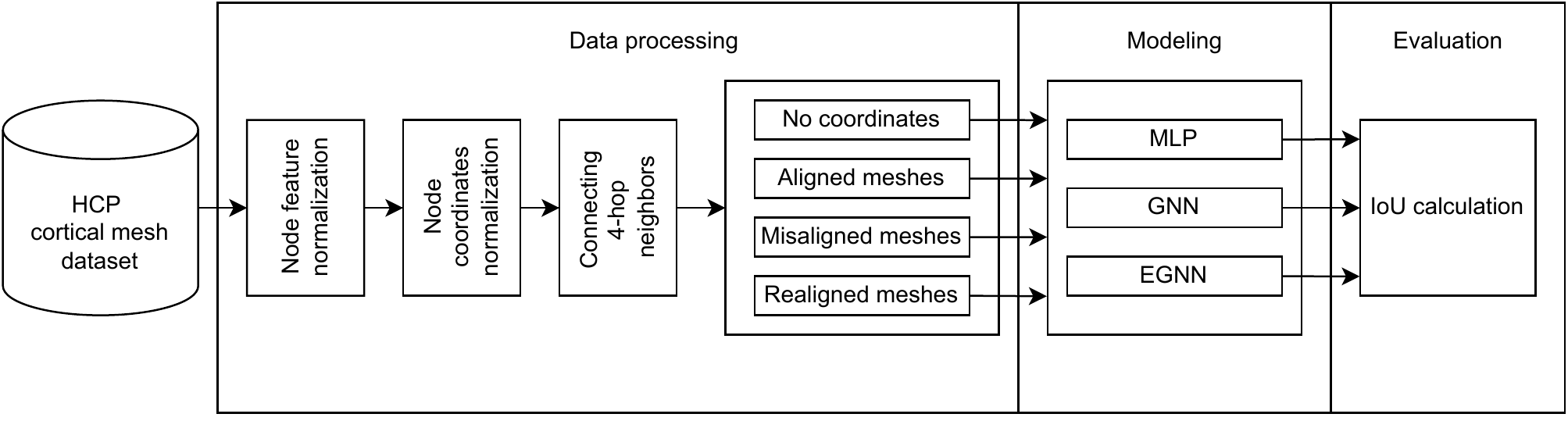}
  \caption{Main steps of the proposed methodology.}
  \label{fig:methodology}
\end{figure}

We show the main steps of our methodology in Figure \ref{fig:methodology}. The first step was data preprocessing, introduced in Subsection \ref{sec:data}, and the four approaches for node coordinate representations in Subsection \ref{sec:experiments}. As the next step, modeling was performed with three distinct neural network architectures: a multilayer perceptron (MLP), a plain graph neural network (GNN) and an E(n)-equivariant graph neural network (EGNN). As the final step, evaluation was performed. 

\subsection{Dataset and preprocessing}
\label{sec:data}
\begin{figure}[!t]
  \centering
  \includegraphics[width=1.0\linewidth]{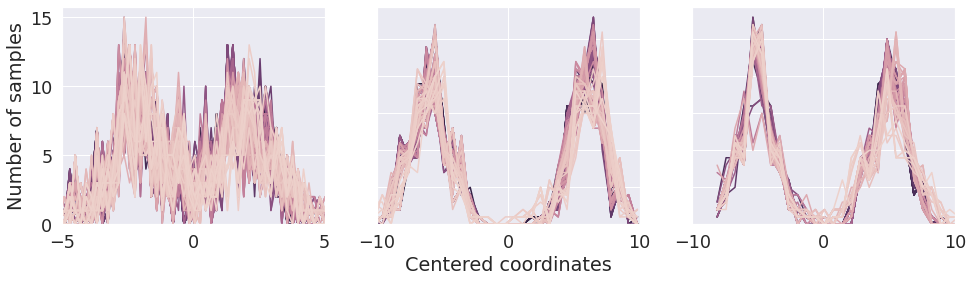}
  \caption{Histograms of the distribution (across all 100 meshes) of node coordinate pairs which lie at a distance that is close to the typical size of the meshes, along each axis. For all pairs, two peaks are distinguishable, as the variability of node coordinates across meshes is much smaller than the node distances. Hence a node can be identified from its position. We plotted the pairs that lie at a distance between 0.98 and 1.02 of the typical size of the meshes, along each axis (calculated as the minimum standard deviation of the given coordinate). 
  }
  \label{fig:data}
  \vspace*{2mm}
\end{figure}

The data we used comes from the \href{http://www.humanconnectomeproject.org/}{Human Connectome Project} (HCP). The dataset consists of 100 cortical meshes, one for each of 100 human subjects. The meshes share the same edge structure, so they are represented by the same adjacency matrix. Each mesh has 1195 nodes, and each node has 6 structural features (cortical thickness, myelin, curvature, sulcal depth, folding corrected cortical thickness and bias-corrected myelin), 3 functional features (rsfMRI correlation with anterior temporal and two parietal regions of interest~\cite{jakobsen}), and the 3-dimensional Cartesian coordinates of the node. Same nodes have approximately the same coordinates across the 100 meshes, as shown in Figure \ref{fig:data}.
Furthermore, each node has a single, manually-assigned label according to the region of the cerebral cortex it belongs to: Brodmann area 44 (BA44), Brodmann area 45 (BA45), both parts of the Broca's area on the left hemisphere of the cerebral cortex, or neither (background)~\cite{jakobsen_label}. The segmentation of this area is particularly challenging as it shows high variability across subjects \cite{jakobsen}.

Data preprocessing involved three steps. First, we normalized the node features to sum-up to one. Second, we centered and normalized the node coordinates to the interval [-1, 1]. Third, we connected the nodes with all other nodes within their 4-hop neighbourhood as it consistently improved the IoU score of GNN and EGNN.

\subsection{Models, training and evaluation}
We applied the following models in our work:

\textbf{Multi-Layer Perceptron (MLP)}: this model is a stack of 6 linear layers, with 32 units in each hidden layer, resulting 5k learnable parameters. We applied ReLU activation and batch normalization after each layer, except the last one where we applied softmax to output label probabilities. 

\textbf{Graph Neural Network (GNN)}: this model has an encoder of one linear layer, 4 hidden message passing layers, and a decoder of one linear layer. The edge and node MLPs have 2 layers, with 32 units in each layer, resulting 26k learnable parameters. We applied Swish activation~\cite{swish} after each layer, except the last one where we applied softmax to output label probabilities.  

\textbf{E(n)-equivariant GNN (EGNN)}: this model has the same architecture as GNN, except it has coordinate MLPs beyond the edge and node MLPs. The coordinate MLPs have 2 layers, with 32 units in each layer, resulting a total of 30k learnable parameters. 

We trained the models in two stages, using the class-averaged dice loss (\(\dice = 1 - 2 \cdot \iou\)) as loss function~\cite{dice}. In the first stage, we used the Adam optimizer~\cite{adam} with learning rate 0.001. 
The first stage was halted when the validation loss did not decrease in the last 200 epochs, and we restored the parameters of the best performing model. In the second stage, we trained the restored model further with SGD + momentum~\cite{sgd_momentum}, where the learning rate and momentum were set to 0.001 and 0.9, respectively. The second stage was halted after 200 epochs, and once again, we restored the parameters of the best performing model. The batch size was set to 10.

All results were obtained by the 10-fold cross validation of the models. We split the data such that we used eight folds for training, one for validation, and the remaining one for testing. We repeated this process 10 times, each time with a different test fold, and report the means and standard deviations of results across the different test folds. Results are reported in terms of Intersection over Union (IoU, also referred to as Jaccard index) of the two classes of interest (BA44 and BA45):
\[ \iou = \frac{y\hat{y}}{y + \hat{y}} \]
where \(y\) are the ground-truth labels (one-hot encoded), and \(\hat{y}\) are the label probabilities predicted by the network.

\subsection{Experiments}
\label{sec:experiments}
\subsubsection{Segmentation using node adjacencies}
\label{sec:data_nocoord}
In our first experiment, we measured the performance of MLP and GNN without using the node coordinates (the column \textit{without coord.} in Table \ref{table:results}). Since MLP cannot exploit node adjacencies, it relies solely on node features, not using any kind of positional information. As a result, its IoU score is well below the IoU score of rule-based methods reported in previous papers~\cite{cucurull,jakobsen}. On the contrary, GNN is able to exploit node adjacencies and outperforms the aforementioned rule-based methods, as explored by Cucurull et al.~\cite{cucurull}.

\begin{figure}[!t]
  \centering
  \vspace*{2mm}
  \includegraphics[width=1.0\linewidth]{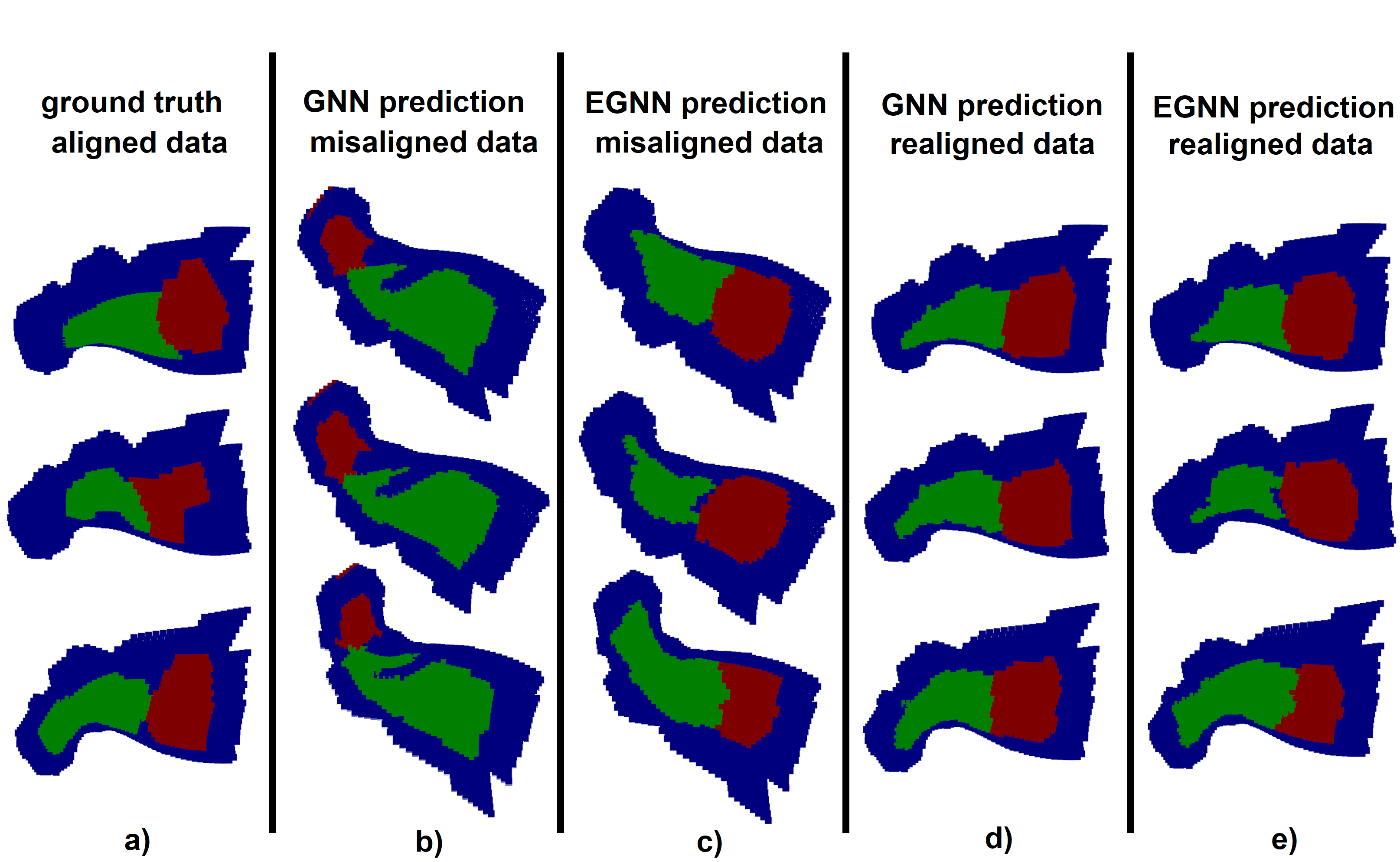}
  \vspace*{2mm}
  \caption{Segmentation results when the test set is misaligned by a rotoreflection. The segmented areas are colour-coded: BA44 by red, BA45 by green, and background by blue. EGNN is unaffected by isometric transformations, producing the same segmentation results for misaligned and realigned meshes (Fig.~3c and Fig.~3e). GNN completely mispredicts the two areas when meshes are misaligned (Fig.~3b), but provides better results than EGNN following realignment (Fig.~3d).}
  \label{fig:segmentation}
\end{figure}

\subsubsection{Segmentation using node coordinates}
\label{sec:data_node}
Next we investigated whether and how using node coordinates helps our models to achieve better results. In MLP and GNN, we simply concatenated the node coordinates to the node features. We also considered the E(n)-equivariant GNN (EGNN) model. In EGNN, we applied the node distances as edge features, and transformed the node coordinates separately (see the Background section for the details).

Based on the results (the column \textit{aligned meshes} in Table \ref{table:results}), we can safely conclude that MLP and GNN perform better than EGNN. The reason is that EGNN is a coordinate-free method: on one hand, its output does not depend on the arbitrary choice of coordinates; but on the other hand, it cannot exploit that our cortical meshes are co-registered in a global coordinate system. In opposition, the output of MLP and GNN depend on the choice of coordinates, and so they can assign a label probability to each coordinate in 3D space. 

\subsubsection{Segmentation following misalignment}
\label{sec:data_mis}
To emphasize the importance of alignment, we generated 100 random isometric transformations, one for each of the cortical meshes. An isometric transformation consists of an orthogonal matrix and a translation vector. Both were generated according to a uniform distribution, and the 3 components of the translation vector were bounded by [-1, 1]. After transforming the meshes, we re-evaluated the performance of the networks. Regarding real-world implications, such misalignment may occur due to the different co-registration protocols, for instance when the data is coming from multiple sources, or a pre-trained model is being used.

The results are in agreement with our previous interpretation (the column \textit{misaligned meshes} in Table \ref{table:results}). Once we misalign the meshes, the performance of MLP and GNN decreases by a significant amount. EGNN performs better than MLP and GNN, producing approximately the same IoU score as in the case of aligned meshes. Because GNN can still rely on the positional information conveyed by node adjacencies, it is considerably more robust to misalignment than MLP.

To further illustrate the importance of alignment, we generated two different isometric transformations, one for the train set and one for the test set. Such transformations severely fool MLPs and GNNs: they need to be evaluated in the same coordinate system they were trained in, otherwise they produce worse test IoU than a randomly generated one. EGNN is free from such constraints: it produced consistent test results as before (illustrated in Figure \ref{fig:segmentation}).

\subsubsection{Segmentation following realignment}
\label{sec:data_re}
As our final experiment, we tried to recover the performance of MLP and GNN by realigning the misaligned cortical meshes. We selected an arbitrary reference mesh (the first sample of the shuffled training set), and estimated the isometric transformation between the reference and the other meshes. Since coordinates that belong to the same node are somewhat different across meshes, we applied the point cloud registration algorithm by Arun et al. \cite{point_cloud} in an iterative manner (20 iterations).

We obtained almost the same result as for the original meshes (the column \textit{realigned meshes} in Table \ref{table:results}). It shows that we successfully co-registered the meshes in a global coordinate system, in which case MLP and GNN can perform better than EGNN. We conclude that the co-registration of meshes (or point clouds) in a global coordinate system and using coordinate dependent methods altogether works better than using coordinate independent methods.

\begin{table}[t]
\caption{Segmentation results on cortical meshes coming from the HCP Dataset, reported in terms of per-class and average IoU across test folds (mean ± standard deviation). }
\centering
\begin{tabular}{p{1.8cm}p{1.8cm}p{1.8cm}p{1.8cm}p{1.8cm}p{1.8cm}}
\toprule
\centering \textbf{Model} &
\centering \begin{tabular}{@{}c@{}}\textbf{Brodmann} \\ \centering \textbf{area}\end{tabular} &
\centering \begin{tabular}{@{}c@{}}\textbf{without} \\ \centering \textbf{coord.}\end{tabular} &
\centering \begin{tabular}{@{}c@{}}\textbf{aligned} \\ \centering \textbf{meshes}\end{tabular} &
\centering \begin{tabular}{@{}c@{}}\textbf{misaligned} \\ \centering \textbf{meshes}\end{tabular} &
\centering \begin{tabular}{@{}c@{}}\textbf{realigned} \\ \centering \textbf{meshes}\end{tabular} \cr
\midrule
& \centering BA44 & \centering 46.9±3.6 & \centering 60.5±3.4 & \centering 42.6±4.7 & \centering 60.8±3.4 \cr
\centering MLP & \centering BA45 & \centering 27.7±5.3 & \centering 52.8±7.1 & \centering 22.2±4.7 & \centering 53.2±7.0 \cr
& \centering average & \centering 37.3±3.3 & \centering 56.6±4.4 & \centering 32.4±3.7 & \centering 57.0±4.4 \cr
\midrule
& \centering BA44 & \centering \textbf{59.3±3.7} & \centering \textbf{61.4±3.1} & \centering 54.3±3.1  & \centering \textbf{61.6±3.6} \cr
\centering GNN  & \centering BA45 & \centering \textbf{51.8±7.9}  & \centering \textbf{53.5±6.9} & \centering 49.1±6.4 & \centering \textbf{53.4±6.8} \cr
& \centering average & \centering \textbf{55.6±4.5} & \centering \textbf{57.4±3.8} & \centering 51.7±3.7 & \centering \textbf{57.5±4.2} \cr
\midrule
& \centering BA44 & \centering - & \centering 60.4±5.3 & \centering \textbf{60.5±4.3} & \centering 59.8±3.9 \cr
\centering EGNN & \centering BA45 & \centering - & \centering 51.7±6.2 & \centering \textbf{51.9±6.0} & \centering 52.1±8.2 \cr
& \centering average & \centering - & \centering 56.1±4.2 & \centering \textbf{56.2±3.5}  & \centering 56.0±4.5 \cr
\bottomrule
\end{tabular}
\label{table:results}
\end{table}

\section{Conclusion}

In this work, we segmented the cortical mesh of 100 human subjects using graph neural networks (GNNs). We demonstrated that co-registering the meshes in a global coordinate system allows GNN to perform better than the isometry-equivariant EGNN. It happens because GNN can exploit the alignment of meshes and assign a label probability to each coordinate in 3D space. We also showed that once we ruin this alignment -- a situation we reproduce via random isometric transformations -- EGNN is unaffected and superior to GNN. These findings are relevant to situations in which training data is not co-registered with test data. For instance, when the data is coming from multiple sources, or a pre-trained model is being used. Co-registration, if possible, eliminates the requirement of E(n)-equivariance in this application domain. Regarding future research, we are planning to focus on intrinsic mesh CNNs~\cite{geometric}, whose prediction depends only on the intrinsic shape of the 2D mesh, and not on its embedding in 3D space.


\section{Acknowledgement}
The authors are especially grateful to Konrad Wagstyl for his valuable insights into the data. The research reported in this paper has been partly supported by the Hungarian National Laboratory of Artificial Intelligence funded by the NRDIO under the auspices of the Hungarian Ministry for Innovation and Technology. We thank for the usage of the ELKH Cloud GPU infrastructure (\url{https://science-cloud.hu/}) that significantly helped us achieve the results published in this paper. We gratefully acknowledge the support of NVIDIA Corporation with the donation of the NVIDIA GPU also used for this research. The publication of the work reported herein has been supported by ETDB at BME.

\end{document}